\documentclass[10pt,twocolumn,letterpaper]{article}

\usepackage{cvpr}              % To produce the CAMERA-READY version
\usepackage{graphicx}
\usepackage{amsfonts}
\usepackage{amssymb}
\usepackage{dsfont}
\usepackage{amsmath,bm}
\usepackage[accsupp]{axessibility}

\usepackage{booktabs}
\usepackage{lipsum}

\usepackage{pifont}% http://ctan.org/pkg/pifont
\usepackage[pagebackref,breaklinks,colorlinks]{hyperref}

\usepackage[capitalize]{cleveref}
\crefname{section}{Sec.}{Secs.}
\Crefname{section}{Section}{Sections}
\Crefname{table}{Table}{Tables}
\crefname{table}{Tab.}{Tabs.}

\def\cvprPaperID{5563} % *** Enter the CVPR Paper ID here
\def\confName{CVPR}
\def\confYear{2022}

\newcommand{\stitle}[1]{\vspace{1ex}\noindent\textup{\textbf{#1}}}

\begin{document}

\title{RendNet: Unified 2D/3D Recognizer with Latent Space Rendering}

\author{Ruoxi Shi\thanks{This work was done when the author was an intern at MSRA.}\\
Shanghai Jiao Tong University\\
{\tt\small eliphat@sjtu.edu.cn}
\and
Xinyang Jiang\\
Microsoft Research Asia\\
{\tt\small xinyangjiang@microsoft.com}
\and
Caihua Shan\\
Microsoft Research Asia\\
{\tt\small caihuashan@microsoft.com}
\and
Yansen Wang\\
Microsoft Research Asia\\
{\tt\small yansenwang@microsoft.com}
\and
Dongsheng Li\\
Microsoft Research Asia\\
{\tt\small dongsheng.li@microsoft.com}
}
\maketitle

%%%%%%%%% ABSTRACT
\begin{abstract}

Vector graphics (VG) have been ubiquitous in our daily life with vast applications in engineering, architecture, designs, etc.
The VG recognition process of most existing methods is to first render the VG into raster graphics (RG) and then conduct recognition based on RG formats. However, this procedure discards the structure of geometries and loses the high resolution of VG. Recently, another category of algorithms is proposed to recognize directly from the original VG format. But it is affected by the topological errors that can be filtered out by RG rendering. Instead of looking at one format, it is a good solution to utilize the formats of VG and RG together to avoid these shortcomings. 
Besides, we argue that the VG-to-RG rendering process is essential to effectively combine VG and RG information. 
By specifying the rules on how to transfer VG primitives to RG pixels, the rendering process depicts the interaction and correlation between VG and RG. 
As a result, we propose RendNet, a unified architecture for recognition on both 2D and 3D scenarios, which considers both VG/RG representations and exploits their interaction by incorporating the VG-to-RG rasterization process.
Experiments show that RendNet can achieve state-of-the-art performance on 2D and 3D object recognition tasks on various VG datasets.

%Synthetic 2D images and 3D scenes have been ubiquitous in our daily life.    They are usually described with a set of primitives. Their inherent arbitrary resolution, as well as the underlying structure of geometries commonly available in this type of data, are largely neglected if we only make use of the rasterized results.    On the other hand, we also notice that the rasterization process imposes prior knowledge about the primitive descriptions since we have no way to know how the attributes affect the outcome without further specification.    Thus, we propose RendNet, a unified architecture for 2D and 3D object recognition. RendNet incorporates a rasterization process that enables it to leverage the different representations of objects well. Primitive descriptions are represented as a hypergraph. For each block in RendNet, there are two flows of streams. In the first stream the primitive descriptions are processed with a hypergraph neural network. In the second stream latent space rasterization is invoked to obtain point clouds. The point-wise features are transformed and aggregated back to the hypergraph node features. Experiments show that RendNet can achieve state-of-the-art performance on 2D and 3D object recognition tasks on synthetic datasets.
\end{abstract}

\newcommand{\placeholder}{PLACEHOLDER. }
\newcommand{\placeholderr}{\placeholder \placeholder}
\newcommand{\placeholderrr}{\placeholderr \placeholderr}
\newcommand{\placeholderrrr}{\placeholderrr \placeholderrr}
\newcommand{\placeholderrrrr}{\placeholderrrr \placeholderrrr}
\newcommand{\placeholderrrrrr}{\placeholderrrrr \placeholderrrrr}

\section{Introduction}

% Do not put the intro figure in the top of left column.
\global\csname @topnum\endcsname 0
\global\csname @botnum\endcsname 0

\begin{figure}[t]
    \centering
    \includegraphics[width=\linewidth]{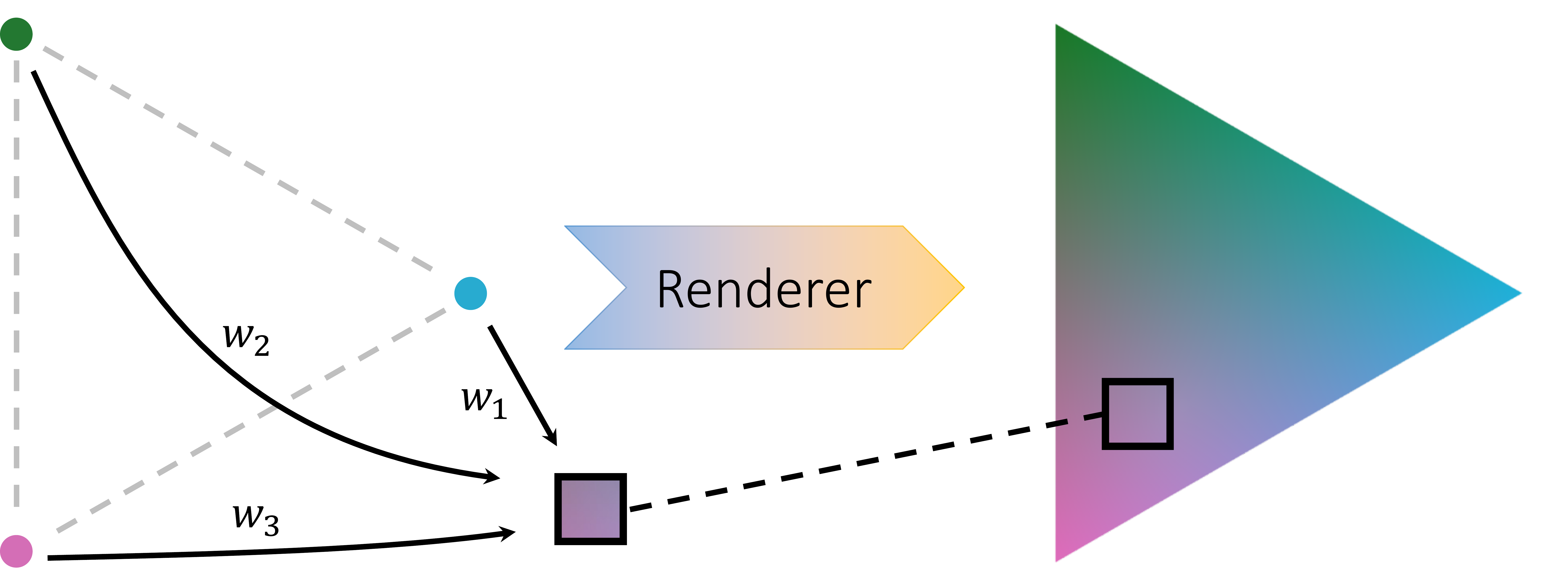}
    \caption{\textbf{Rendering of vector graphics.} The renderer knows the correlation between sparse VG attributes and the rendered RG.}
    \label{fig:intro}
\end{figure}

Deep learning has opened a new era for visual perception with machines. Most current methods deal with sensory input such as pixel images, called raster graphics (RG). They benefit from the easy accessibility of input data. 
However, for human drafted graphics like floor plans, graphic designs, and CAD models, another data format is widely used, called vector graphics (VG). In this paper, we focus on recognition tasks taking VG as input, such as VG-based image classification and object detection. 

\begin{figure}[t]
    \centering
    \includegraphics[width=0.63\linewidth]{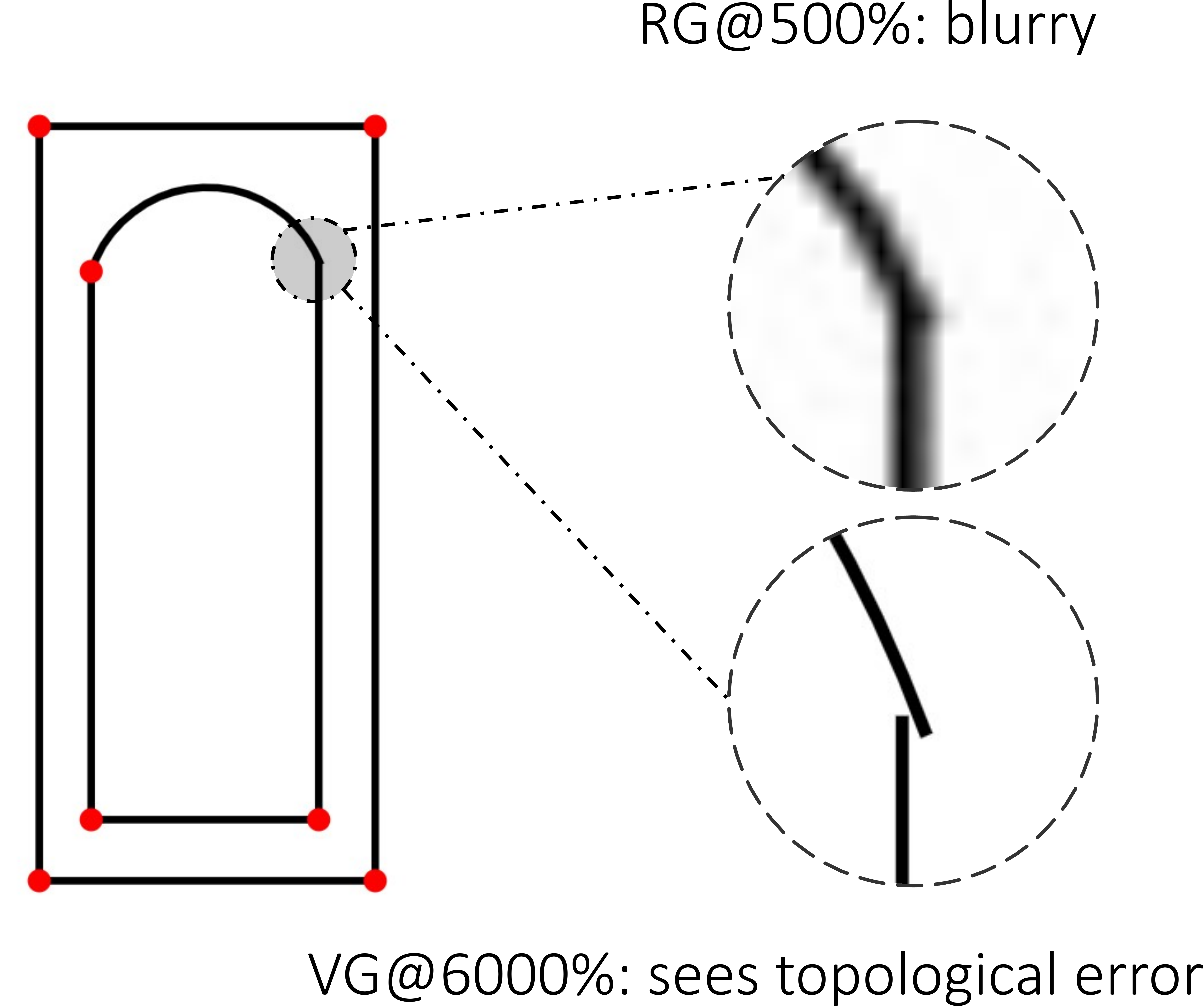}
    \caption{\textbf{Cons of different graphics formats}. A comparison between different representations of a SESYD floor plan symbol instance. The raster graphics get blurry, or aliased, if we zoom in. The vector graphics stay clear. However, it has a topological error. This adds noise in VG data.}
    \label{fig:rvg_proscons}
\end{figure}

Vector graphics contain a set of primitives defined with parametric equations, such as lines, curves, and circles, which are almost impossible for a human to directly perceive. 
%When we operate with data rendered into sensory format, the structure of geometries and arbitrary resolution of vector graphics are largely neglected. 
It needs to be rendered into the format as the raster graphics by rasterization technique (as shown in Fig.~\ref{fig:intro}), so it can be displayed on monitors or printed on paper. 
Most of the existing VG recognizers take the rendered RG as input, taking advantage of mature RG-based recognition methods like convolutional neural networks \cite{he2016deep} or PointNet \cite{qi2017pointnet}. 
However, rendered pixels discard the structure of geometries and lose the high-resolution property of VG. 
As a result, recently some pioneering works \cite{jiang2021recognizing,uvnet} are proposed to directly recognize VG from its original format. 
Although achieving encouraging performance improvements, VG-based methods are affected by human un-perceivable topological errors, which can be filtered out by rasterization.
%This is because when human designers draft VG, they do it by looking at the rendered RG, and hence . 
For example,  Fig.~\ref{fig:rvg_proscons} shows two lines whose ends shall meet but do not meet by a small margin, and this error is absent in the rendered RG. 
Different from the existing methods that consider only one format, this paper proposes a method that leverages the merits of both VG and RG. 

%VG and RG have their own merits and disadvantages. On one hand, VG is more compact, containing more visual information than RG.Described with a set of analytical represented primitives, vector graphics have arbitrary resolution, structure of  geometries, and sparse representations. On the other hand however, RG is a human perceivable format, and when human designers draft VG, they do it by looking at rasterized images. As a result,  there may be topological errors in VG (see Fig.~\ref{fig:rvg_proscons} where two line ends shall meet but do not meet by a small margin) that is invisible to human on raster graphics. In other words, RG can filter out human un-perceivable noises from the original vector format. Existing methods only focus on one of the two formats \cite{jiang2021recognizing,brepnet}, while this paper proposes to leverage merits of both RG and VG. 
% Furthermore, due to fabricating limits, it is common to break down a larger and more complex primitive into smaller and simpler ones (see Fig.~\ref{fig:rvg_proscons}b). These bring non-uniqueness, or ambiguity, to the raw data: the same object in design may be expressed by many different representations in VG; there lack a canonical representation of objects in VG.
% {\color{red} comments from Dongsheng: I don't quite understand this paragraph. What will the topological errors cause as in Fig 2a? What kind of ambiguity will be brought by the example in Fig 2b?}

How to effectively combine the information of VG and RG remains an open question. 
An intuitive way is to use the separate model on VG and RG respectively and fuse the results.
However, it fails to exploit the interactions and correlations between the VG primitives and RG pixels, which is essential to multi-modality feature fusion \cite{baltruvsaitis2018multimodal,wang2020incorporating,zadeh2017tensor}. 
As shown in Fig.~\ref{fig:intro}, the VG-to-RG renderer reveals VG/RG correlation by specifying the rules on how to transfer VG primitives to RG pixels. 
As a result, we consider incorporating the rendering process as a part of our method to better model the interactions and correlations between RG and VG.

In view of these problems, we propose a novel vector graphics recognition method, named RendNet.  RendNet leverages both VG/RG representations and exploits their interactions by incorporating the rendering process.
%It is a unified framework applicable on both 2D (e.g., SVG) and 3D (e.g., BReps) VG data. Specifically, RendNet takes the VG primitives as input and represent the VG as a hypergraph. The hypergraph is fed into a network with two parallel streams: a `vector' stream that handles hypergraph topology and extracts VG embeddings, and a `raster' stream that extracts spatial embeddings of RG. 
It is a unified framework applicable to both 2D and 3D scenarios. 
Specifically, RendNet takes the VG primitives as input and represents the VG as a hypergraph. The hypergraph is fed into a network with two parallel streams. A `vector' stream handles hypergraph topology and extracts VG embeddings by hypergraph neural networks. A `raster' stream transfers the VG into 2D pixels or 3D point clouds as RG information. To unify the framework, we utilize PointNet as a simple but effective backbone to extract spatial embeddings of RG.
Finally, as the core of RendNet, the correlation between VG and RG embeddings is modeled by a novel latent space rasterization (LSR) method,  which simulates a rendering process and projects the VG representations to RG latent features. In this way, the VG/RG correlations are exploited at multiple semantic levels throughout the entire network.

%It projects VG primitives represented by node features on a hyper-graph  onto a dense RG point cloud. %, which is further illustrated in Sec.~\ref{sec:fsr}. 

 %By incorporating rasterization into the model pipeline, VG and RG are integrated closely with each other. The model can thus leverage the benefits from both formats.
%Specifically, RendNet involves multiple residual blocks and a final block to aggregate global features. The main features that flow between blocks are the node features on the hypergraph. For each residual block, the input goes through two parallel streams: a `vector' stream that handles hypergraph topology and features; a `raster' stream that invokes LSR, after which a PointNet is applied to Euclidean neighbourhoods of hypergraph nodes to obtain new hypergraph node features. 

% For evaluation, we benchmark RendNet against previous methods for various object recognition tasks. For 2D image classification and object detection, it is evaluated on the SESYD \cite{delalandrerecent} floor plans and diagrams datasets. For 3D tasks, we evaluate it on the machining feature classification dataset from FeatureNet \cite{zhang2018featurenet} and the more unconventional similarity assignment Cluster3D \cite{xiang2021cluster3d} dataset. RendNet achieves state-of-the-art performances on these tasks.

\stitle{Our main contributions:}
1) RendNet is a 2D/3D unified vector graphics recognition framework that leverages the merits of both vector graphics and raster graphics. 
2) RendNet incorporates the rendering process and effectively exploits the interaction between RG and VG. 
3) Experiments are conducted on both classification and object detection tasks on both 2D and 3D datasets. State-of-the-art performances are achieved. 

\section{Related work}
\stitle{2D VG recognition and generation. } 
Most of the concurrent recognition methods focus on pixel-based raster graphics, such as ResNet\cite{he2016deep}, R-CNN\cite{girshick2014rich,girshick2015fast,ren2015faster}, YOLO \cite{redmon2016you,redmon2017yolo9000}. Thus, the intuitive way to conduct VG recognition is a two-stage process that applies CNN-based models on rendered VGs. 
One of the applications of VG recognition is architecture drawing recognition. Several rule-based graph matching methods are  proposed to classify and localize symbols by representing symbols in a floor-plan as graphs for matching, such as visibility graph \cite{locteau2007symbol} and attributed relational graph \cite{ramel2000structural,santosh2012symbol}. 
Online handwriting \cite{ha2017neural,NEURIPS2020_723e8f97} is a data form similar to vector graphics. Most of these methods use sequential models to handle the point sequence in hand-writing. Compare to online handwriting, VG contains more types of primitives and has a more general and complex topological structure. 
Another AI application on VG is computer-aided design. 
In recent years, a few works propose to use deep learning to automatically generate design graphics or convert raster graphics to vector graphics (i.e. vectorization) \cite{DBLP:conf/nips/CarlierDAT20,ganin2021computer,reddy2021im2vec,lopes2019learned, shen2021clipgen, parakkat2021color}. %Koch~et~al. \cite{koch2019abc} proposes a large 3D model dataset containing analytic representations, but it lacks semantic labeling to train recognition model. 

\begin{figure*}[!t]
    \centering
    \includegraphics[width=0.85\linewidth]{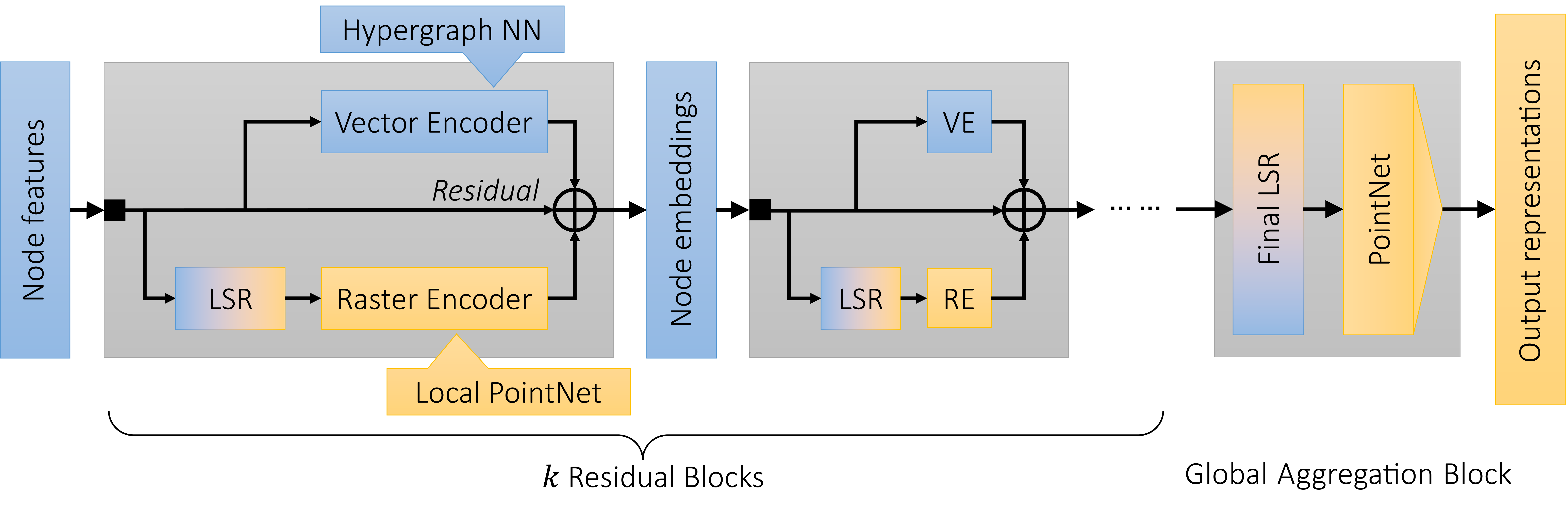}
    
    \caption{\textbf{Overview of RendNet architecture.} RendNet is made by $k$ two-stream residual blocks, followed by a final block for global representation aggregation. The output representation can be fed into downstream tasks.}
    \label{fig:raster_gnn_arch}
\end{figure*}
\begin{figure*}[!t]
    \centering
    \includegraphics[width=0.9\linewidth]{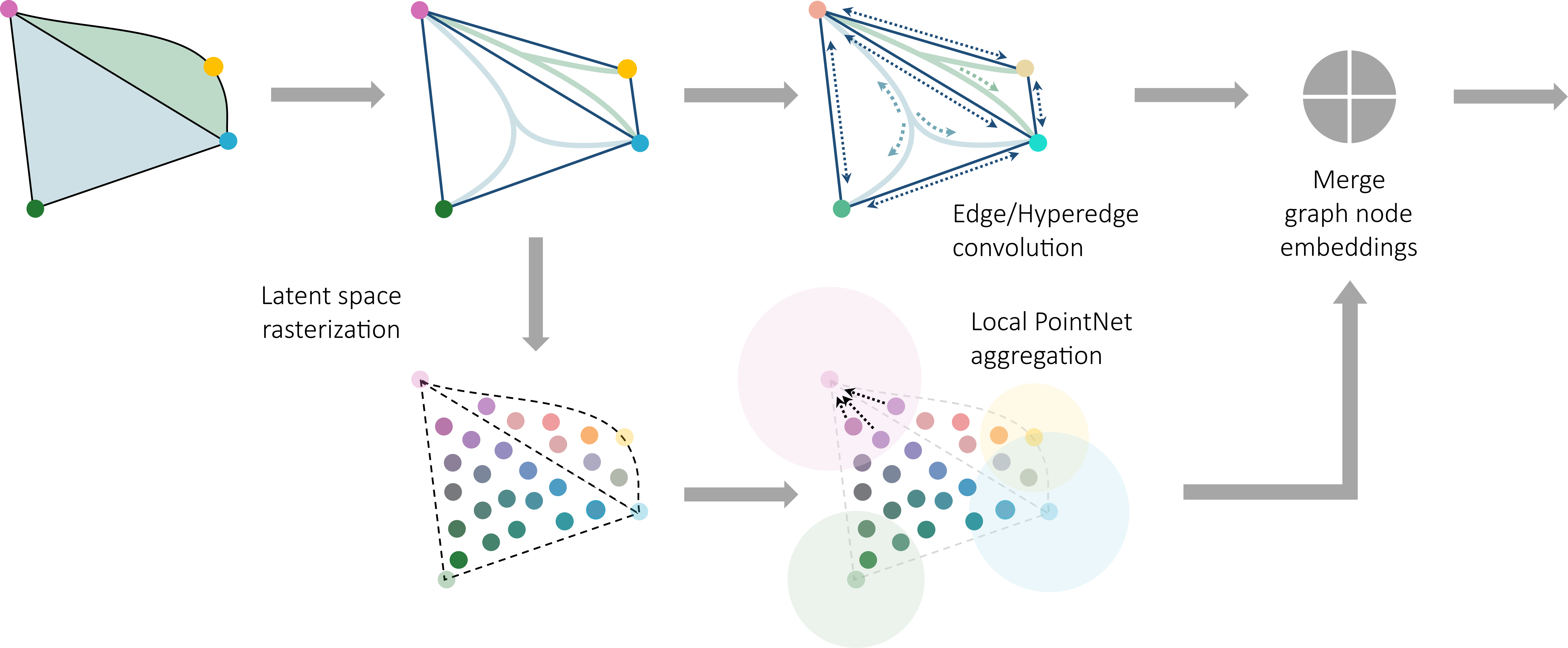}
    
    \caption{\textbf{A residual block in RendNet.} It is made of a vector stream and a raster stream. Their output latent representations are merged by a summation. Latent feature values are indicated by colors.}
    \label{fig:resblk}
\end{figure*}

\stitle{3D representations.} 3D objects can be represented with different formats including point clouds, voxels, meshes, multi-view images, boundary representations (BRep). 
% Among all, point clouds are the most commonly used data format.
Many deep learning techniques have been studied to address various problems on point clouds, such as shape classification, object detection and segmentation~\cite{3dsurvey}. PointNet~\cite{qi2017pointnet} and PointNet++~\cite{qi2017pointnet++} are two pioneering works which capture fine-grained geometric information of each point. PointConv~\cite{wu2019pointconv} defines convolutional kernels and takes a local subset of points as inputs to learn the hierarchical relations. DGCNN~\cite{wang2019dynamic} constructs a graph where the vertex is each point in a point cloud and the edge is generated based on the neighbors of each point. % Then EdgeConv and max pooling are used as the graph convolution and aggregation function to dynamically update the embedding of each point from its neighbors.
Further, volume-based methods~\cite{maturana2015voxnet,zhou2018voxelnet} usually quantize point clouds or other 3D formats into voxels and utilize 3D convolutional neural networks to learn representations upon. Besides, multi-view methods~\cite{su2015multi, wei2020view} project a 3D shape into multiple 2D views, extract view-wise features, and aggregate into a global representation. 

Different from the previous four formats, the boundary representation (BRep), widely used during the CAD modeling process, is made of a variety of parametric primitives such as Bézier curves and planes. It is a typical `VG' data in 3D. Very recently, UV-net and BRepNet \cite{uvnet,brepnet} are proposed to directly utilize BRep topological information by constructing a graph based on face-edge relations. 
%They do not transfer BReps into point clouds or other formats.
%and they perform well on face classification and segmentation.
% They perform well on face classification and segmentation. 
In contrast to these works, we leverage the merits of two formats, BReps and point clouds by rasterization technique. % We learn precise representations of edges and surfaces through B-rep, render them into point representations, and utilize PointNet to predict with large error tolerance. 

\stitle{Differentiable rendering.} Differentiable rendering involves a rendering process where gradients of object attributes can be propagated through the rendering results \cite{kato2020differentiable}. Recent works \cite{pavlakos2018learning,genova2018unsupervised,bao2019high,rhodin2015versatile} utilize differentiable rendering to form an unsupervised workflow for human pose detection and face reconstruction, based on consistency between the original image and the rendered image from detected features. In this work, we choose to render the VG primitives in the latent space rather than the pixel space. The rendering target is a point cloud with latent feature attributes, instead of color attributes in normal rendering.

\section{Design of RendNet}

\subsection{Overview}

In this section, we describe the framework of our proposed RendNet and then introduce each module in detail. As shown in Fig.~\ref{fig:raster_gnn_arch}, it is composed of $k$ basic blocks with residual connections and a final block for global representation aggregation. 

The original input is a set of primitives. We first convert them into a hypergraph with the initial node features $h_i^{0}$ based on Sec.~\ref{sec:build_hypergraph}. Then we learn the hypergraph representation by $k$ residual blocks. Inside each block, node embeddings pass through two streams, the vector stream and the raster stream, to perform the graph convolution and point representation aggregation to obtain the new node embeddings $h_i^{l+1}$. Fig.~\ref{fig:resblk} shows the detailed process of two streams: the vector stream includes vector encoder (Sec.~\ref{sec:vector_encoder}), which utilizes hypergraph neural networks (hypergraph NN) to aggregate node embeddings mainly based on the connectivity (i.e., nodes connected by curve segment or falling on the same surface); the raster stream involves latent space rasterization (Sec.~\ref{sec:fsr}) and raster encoder (Sec.~\ref{sec:raster_encoder}), which renders VG node embeddings as 2D pixels or 3D point clouds and learn the representation in the Euclidean space. We integrate the information from two streams with residual connection to get the new embeddings (Sec.~\ref{sec:integrate_two_encoders}).

Especially, we highlight the LSR operation because it is the bridge between VG and RG. Through LSR and residual connection among blocks, the information can be propagated 
in various paths, such as VG$\to$LSR$\to$RG, LSR$\to$RG$\to$VG, VG$\to$LSR$\to$RG$\to$VG, etc. Finally, the global aggregation block  (Sec.~\ref{sec:global_feature_agg}) learns the final representation for the whole hypergraph, i.e., all the primitives.

%We will discuss the design of the modules in detail in separate sections below.
%Now we introduce the architecture of proposed RendNet. It is composed of multiple basic blocks with residual connections and a final block for global representation aggregation, as is shown in Fig.~\ref{fig:raster_gnn_arch}. 

%The inputs to each residual block are the current node embeddings on the hypergraph. At the beginning of each residual block, the input goes through a batch normalization process \cite{ioffe2015batch} and a ReLU activation, and then two streams shown in Fig.~\ref{fig:resblk}. The `vector' stream is to handle structural vector graphics information, and the `raster' stream involves latent space rasterization (LSR), learning on the resulting point cloud, and aggregation back to the hypergraph nodes. At the output the streams are summed into the input to the residual block. We highlight the LSR operation first since it is essential for the integration of VG and RG in RendNet.
%\subsection{Hypergraph NN for vector graphics}
\subsection{Building the hypergraph}
\label{sec:build_hypergraph}

In RendNet, we convert all the primitive descriptions in the vector graphics into a hypergraph to learn latent features. The hypergraph is mainly formed by connectivity and entails the property of each primitive and the topology among primitives at the same time.

\stitle{Selection of nodes.} The intersection of curves, as well as the start and end of curves, are picked as nodes. In addition, if the curve is not a straight line, at least 4 nodes are ensured to be selected on it with farthest point sampling. The node features are coordinates of these points in Euclidean space.

\stitle{Curves as edges.} The curve connectivity of the nodes above naturally forms edges between the nodes. Each curve is broken into segments by nodes, and each segment is formulated as an edge. The edge features include the start direction vector, the end direction vector, and the type of the curve. The feature is selected due to the universality between different types of curves. Also, in most current specifications of vector graphics (such as SVG), all types of curves as lines, arcs, and second-order bezier curve segments, can be uniquely reconstructed from these attributes, up to translational and scaling invariances.

\stitle{Surfaces as hyperedges.} Surfaces connect multiple nodes, and thus they form hyperedges on the hypergraph. The hyperedge features include the type of the surface and the set of parameters of nodes lying on the surface. This encodes the relative positioning of nodes on the surface, which is useful for feature aggregation.

\subsection{Vector encoder}
\label{sec:vector_encoder}

The vector encoder aggregates the node/edge/hyperedge features and the hypergraph structure. To achieve it, we utilize hypergraph neural networks to learn node embeddings.

For curve edges, we use the NNConv scheme proposed by Gilmer \etal \cite{gilmer2017neural}:
% \begin{equation}
%     h_{i}^{(l+1)} = \mathrm{mean}\left(\left\{
% \mathbf{f}_\Theta^{(l)} (e_{ij}) \cdot h_j^{(l)}, j\in \mathbb{N}(i) \right\}\right),
% \end{equation}
\begin{equation} 
\label{eq:con_curve}
    C_{i}^{l} = \frac{1}{|\mathcal{N}(i)|} \sum_{j\in \mathcal{N}(i)} {
    \left( h_j^{l} \cdot \mathbf{f_{\Theta_1}} (e_{ij})\right)},
\end{equation}
where $\mathcal{N}(i)$ is the set of neighbors of node $i$. 
%$h_{j}^{l}$ denotes the embedding of node $j$ from the previous $l$-th block. 
We apply $\mathbf{f_{\Theta_1}}$ as an MLP to map edge features to a coefficient matrix, and then multiply the embeddings of neighbors $h_{j}^{l}$ from the previous $l$-th block. 
%, which is multiplied by embeddings of neighbors. 
We average the information from neighbors and obtain the curve message $C_{i}^{l}$.

%We average the message from neighbors and add the $l$-th embedding to obtain the final embedding  $h_{i}^{l+1}$.

For surface hyperedges, we adopt the hypergraph message passing design from Feng \etal\cite{feng2019hypergraph}. The hypergraph message passing consists of two stages, aggregation of connected node embeddings of hyperedges into hyperedge representations, and aggregation of hyperedge representations back to nodes. We improve the message passing scheme with geometry taken into account.

In the first stage, we aggregate the node embeddings in a PointNet \cite{qi2017pointnet} fashion, \ie we first concatenate the $l$-th node embeddings with the relative coordinates on the surface in parameter space, then apply the max aggregation to obtain hyperedge-wise representations:

\begin{equation} 
    h_S^{l} = \max_{i \in S} \mathbf{f_{\Theta_2}}\left(\mathrm{concat}\left[ h_i^{l}, {T}_{S}, {t}_{i,S} \right] \right),
\end{equation}
where $S$ denotes a surface hyperedge and $h_i^{l}$ is the embedding of node $i$ that lies on S. We concatenate it with ${T}_S$, the type of the surface (rectangle/circle, etc.) and ${t}_{i,S}$, the coordinates of node $i$ on $S$. Through an MLP $\mathbf{f_{\Theta_2}}$ and a max aggregation, we obtain the representation $h_S^{l}$ of $S$. Note that the coordinates for the same node are relative and can be different on different surfaces.

In the second stage, we calculate the hyperedge message $D_i^{l}$ for each node by averaging hyperedge representations:
% divided by the number of members of each hyperedge:
\begin{equation} 
\label{eq:con_hyperedge}
    D_i^{l} = {\frac{1}{|S|}\sum_{\left\{ S \mid i \in S \right\}} h_S^{l}}.
\end{equation}
%The two kinds of node embeddings, one from graph convolution on curve edges and one from message passing on surface hyperedges, is summed together as output. 

\subsection{Latent space rasterization}
\label{sec:fsr}

\begin{figure}
    \centering
    \includegraphics[width=0.95\linewidth]{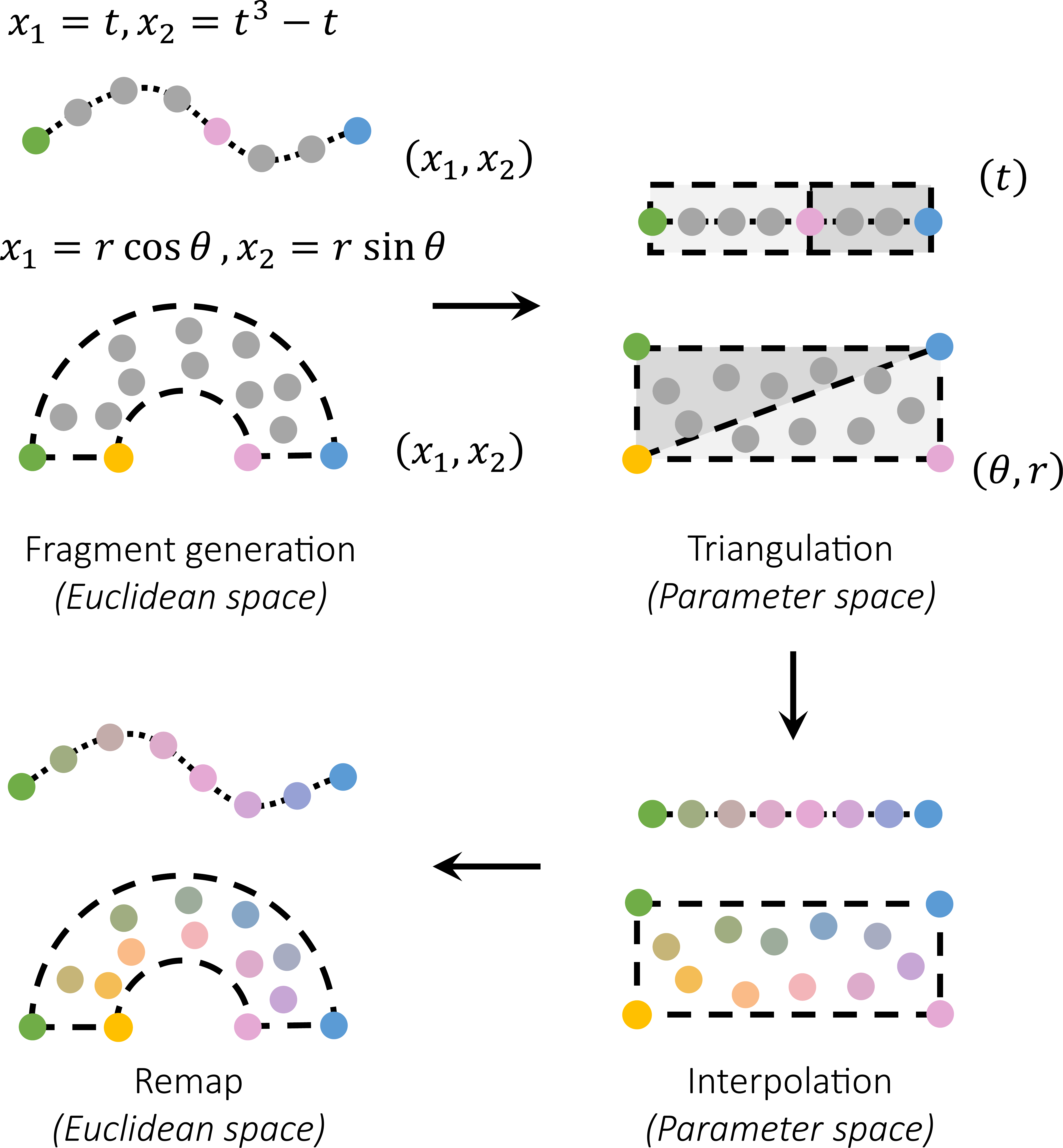}
    \caption{\textbf{Latent space rasterization.} Fragment generation is the first step and the latter three operations form the complete varying interpolation process.}
    \label{fig:lsra}
\end{figure}

In RendNet, we employ a tailored rasterization process that follows the general pattern of rasterization in computer graphics (CG) but is specially designed for incorporating in object recognition, namely the Latent Space Rasterization (LSR). It serves as a renderer from VG to point clouds that operates in latent space. The same process is performed for both 2D and 3D objects. Fig.~\ref{fig:lsra} shows the workflow of LSR. % Fig.~\ref{fig:trirast} shows a comparison between rasterization in CG and LSR. % , but it is not of interest here.

% \sch{cannot understand the meaning of this paragraph. Why not change the order of these two paragraphs? We firstly say "In this section, we do..." and then speacify the interpolation.}
\stitle{Rasterization in CG.} Storing dense samples of attributes in the object space is inefficient and does not support arbitrary-precision rendering of objects. In modern polygon-based computer graphics, triangle meshes are used as primitives for rendering. The attributes are sparsely specified, only on vertices in these meshes. During rasterization, two tasks, \ie \textit{fragment generation} and \textit{varying interpolation}, are done by the rasterizer. Firstly, it projects triangle meshes onto the screen and finds every fragment that falls in the projection region. A fragment is typically a pixel or a sub-pixel region in a multi-sampling scenario. After that, the rasterizer interpolates the sparse attributes on vertices into attributes of each pixel. %Specifically, this is termed varying interpolation. 
We adapt these two tasks in LSR to make them more effective for object recognition but still efficient.

\stitle{Fragment generation.} Since we are not really rendering objects onto the screen, we abandon the projection operation. However, using a pixel/voxel grid as an array of fragments is computationally expensive since the latent space has many more dimensions than the color space. Hence we use point clouds instead of pixels or voxels. 

The primitives (curves and surfaces) in vector graphics are sampled to produce dense point clouds representing raster graphics. For each curve, we sample with a high resolution at equal arc lengths. For each surface, we perform an approximate Poisson disk sampling \cite{yuksel2015sample} so that each point has the same distance to its nearest neighbor. This results in a uniformly distributed point cloud. The number of points on a surface is proportional to the area of the surface.

\stitle{Varying interpolation.} Here we equip each point in the sampled point cloud with a set of latent features from the vertices in the vector graphics.

% \begin{figure}[t]
%     \centering
%     \includegraphics[width=0.85\linewidth]{figs/trirast.pdf}
%     \caption{\textbf{Rasterization in computer graphics and in RendNet.} (a) The color attributes of vertices of a triangle mesh are interpolated into dense pixels when getting rasterized. (b) A door symbol in SESYD floor plans. RendNet rasterizes the sparse representations on the hypergraph vertices into dense point clouds via LSR. The interpolation factors are depicted as colors.}
%     \label{fig:trirast}
% \end{figure}

In computer graphics, in order to shade pixels with correct attributes, we need to interpolate the sparse attributes on triangle vertices. Linear interpolation on triangles is used since three vertices define an affine function of attributes in the triangle nicely.

For arbitrary geometry, we represent the geometry with an array of parametric equations $f_1, f_2, \dots$, where variables $t_1, t_2, \dots$ ranging from $0$ to $1$ form the parameter space:
% \begin{equation}%\small
% \begin{dcases}
% x_1 = f_1(t_1, t_2, \dots)\\
% x_2 = f_2(t_1, t_2, \dots)\\
% \dots
% \end{dcases}
% .
% \end{equation}
\begin{equation} %\small
x_1 = f_1(t_1, t_2, \dots), \quad x_2 = f_2(t_1, t_2, \dots), \quad \dots
\end{equation}
We then invoke a triangulation process to generate simplices of these variables. This is trivial for 1D curves. For higher dimensions, we run a Delaunay triangulation algorithm \cite{lee1980two}. The Delaunay triangulation ensures that no vertex is inside the circumsphere of any other simplex other than the one it belongs to in the triangulation result. Therefore, it will not generate flat simplices that are ill-conditioned for further numerical computations, since flat simplices will have extremely large circumspheres.

Finally, we do linear interpolation inside these simplices. This results in an almost everywhere differentiable attribute function in the object space and thus is a reasonable way of interpolation. More importantly, the result is also differentiable with respect to the sparse attributes from inputs, and thus we may perform backpropagation as usual.

The triangulation is done at the pre-processing stage. The interpolation can be efficiently implemented on GPUs since the process is similar to the special case of varying interpolation in pixel space in modern rendering pipelines.

\subsection{Raster encoder}
\label{sec:raster_encoder}
Based on the latent space rasterization described previously, we can transform vertices in the hypergraph into a dense point cloud. Accordingly, the embedding of each point  $\hat{h}_p^{l}$ is also interpolated from node embeddings $h_i^l$. 

% a kind of point convolution, which
The raster encoder takes point embeddings as input and aggregates the local information in the Euclidean space. 
%It takes in the rendered point cloud from the latent space rasterization described previously, and aggregates the information back into hypergraph node features.
In detail, we first collect the Euclidean $k$-nearest neighbors ($k$-NN) in the rasterized point cloud for each node $i$ in the hypergraph, termed as $\mathcal{M}(i)$. Then a PointNet is utilized to aggregate the point embeddings  back to the hypergraph representation:
\begin{equation}
    \label{eq:raster_agg}
    {E}_i^{l} = \max_{p \in \mathcal{M}(i)} \mathbf{g_\Theta}\left(\mathrm{concat}\left[ \hat{h}_p^{l}, {x}_{p} - {x}_{i} \right] \right),
\end{equation}
where 
%$\mathcal{M}(i)$ is the Euclidean $k$-NN neighbourhood (k nearest points in the rasterized point cloud) of node $i$. 
%$\hat{h}_p^{l}$ is the embedding on point $p$ obtained from latent space rasterization.
${x}_{i}$ and ${x}_{p}$ denote position coordinates of the nodes and the points, respectively. $\mathbf{g_\Theta}$ is an MLP. The use of the $\max$ operation comes from PointNet \cite{qi2017pointnet}, since it is invariant to the permutation of points in a point cloud, and is also agnostic to local point sampling density. 

\subsection{Merging node embeddings}
\label{sec:integrate_two_encoders}
At the end of each block, we sum up all the node messages with the shortcut connection: two from graph message passing on curve edges (denoted as $C_i^l$, Eq.~\ref{eq:con_curve}) and surface hyperedges (denoted as $D_i^l$, Eq.~\ref{eq:con_hyperedge}) in vector encoder and one from PointNet in the Euclidean space (denoted as $E_i^l$, Eq.~\ref{eq:raster_agg}) in raster encoder, as:

\begin{equation}
h_i^{l+1} =  h_i^{l} + {C}_i^{l} + {D}_i^{l} + {E}_i^{l}.
\end{equation}
Before the new node embeddings $h_i^{l+1}$ are passed into the next block, we perform relu activation and batch normalization as \emph{pre-activation}\cite{he2016identity}. % Note that we also apply the pre-activation for the initial node feature $h_i^{0}$. 

\begin{table*}[t]
\small
\centering

\begin{tabular}{ccccc}
\toprule
{Method}& {Data format} & {mAP@$.5$} & {mAP@$.75$}  &  {mAP@$[.5, .95]$}\\
\midrule
{Yolov3-tiny} & RG (pixels) & 75.23 & 60.97 & 53.24\\
{Yolov3} & RG (pixels) &  88.24 & 80.44 & 72.98 \\
{Yolov3-spp}  & RG (pixels) & 87.38 & 79.66 & 71.61  \\
{Yolov4}  & RG (pixels)  & 93.04 & 87.48 & 79.59  \\
{Faster-RCNN-R18-FPN}  & RG (pixels)  & 80.91 & 71.48 & 67.32 \\
{Faster-RCNN-R34-FPN} & RG (pixels) & 80.50  & 72.18 & 65.89  \\
% \textbf{faster-rcnn-R50-FPN} &  &  &     & 73  &   \\
% train with 6x
% \textbf{faster-rcnn-R50-FPN} & \xmark & 85.59 & 78.42 & 70.07 &   &  &   \\
{Faster-RCNN-R50-FPN} & RG (pixels) & 80.31 & 73.28 & 66.53 \\
% train with 1x
% \textbf{retinanet-R50-FPN} & \xmark &  63.65\% & 55.60\% & 49.86\% & 72.2 & 38.0 & 189.2  \\
% train with 3x
{RetinaNet-R50-FPN} & RG (pixels) &  87.50 & 82.91 & 79.18  \\
\midrule
{YOLaT} & VG & \textbf{98.83} & 94.65 & 90.59 \\
\midrule
{RendNet~(Ours)} & VG + RG (point cloud) & 98.70 & \textbf{98.25} & \textbf{91.37} \\
% \midrule
% {faster-rcnn-R50-FPN} & RG & \cmark & 98.04 & 95.23 & 90.25  \\
% {Yolov3} & RG &\cmark & 74.61 & 60.33 & 53.76  \\
% \textbf{retinanet-R50-FPN (pretrained)} & 94.95\% & 86.77\% & 79.63\% & 77.1 & 37.9 & 189.0 \\
% & \textbf{Detr} \\
% \midrule

%\multicolumn{1}{c}{\multirow{2 }{*}{\begin{tabular}[c]{@{}c@{}}
%\textbf{Graph-based} \\ \textbf{models}  \end{tabular}}}&
%\textbf{GCN} &  0.9036 & 0.8802 & 0.8332 & & & 1.6014 & 5.5\\
%&\textbf{GAT} & 0.9120 & 0.8946 & 0.8392 & & & 1.6019 & 5.5\\
%& \textbf{GraphSage} & 0.9270 & 0.9117 & 0.8526 & & & 1.6103 & 5.7\\
%\midrule

% \textbf{\Model~(Ours)} & \xmark & \textbf{98.20\%} & 94.65\% & \textbf{90.47\%}  & 1.3 & \textbf{1.6} & \textbf{1.6}\\
\bottomrule
\end{tabular}
\caption{
\textbf{Performance comparison on SESYD-floorplan in terms of mAP (\%) at different IoU.} RendNet outperforms all the baselines in terms of mAP@$.75$ and mAP@$[.5, .95]$, while achieving comparable performance with YOLaT in terms of mAP@$.5$. 
}\label{table:results-floorplan}
\end{table*}

\subsection{Global feature aggregation}
\label{sec:global_feature_agg}
The structure of the final block for global feature aggregation is similar to the raster stream in the previous blocks, but the whole point cloud is rendered and processed instead of local neighborhoods of nodes. We invoke latent space rasterization of the object and then apply a PointNet to gather global information of the object. In other words, we let the output global representation $h'$ be

\begin{equation}
    h' = \max_{p} \mathbf{g'_\Theta} \left(\mathrm{concat}\left[ \hat{h}_p^{(l)}, {x}_{p} \right] \right).
\end{equation}

This is similar to Eq.~\ref{eq:raster_agg} except that the coordinates and the maximum aggregation function are taken in a global rather than local sense.

\section{Experiments}

In this section, we evaluate RendNet on different tasks and datasets to examine the effectiveness of RendNet on 2D and 3D vector graphics recognition problems. RendNet is implemented with PyTorch \cite{paszke2017automatic} and DGL \cite{wang2019deep}, and the code is provided in the supplementary. 

%\subsection{Object detection on SESYD floor plans}
\subsection{2D object detection}

In this section we evaluate RendNet for 2D object detection on SESYD floor plans \cite{delalandrerecent},
%Following the experimental settings from YOLaT \cite{jiang2021recognizing}, different methods are evaluated on SESYD-floorplan \cite{delalandrerecent}. 
a public dataset of floor plans that have VG sources available. It contains 1000 images, with a total of 28065 objects in 16 categories, like armchairs and windows. %The size of each image is 28,065.

\stitle{Experimental setting.} Following the setting of YOLaT~\cite{jiang2021recognizing}, the images are evenly distributed in 10 layouts. 
%There are Diagrams 16 types of furniture included for detection, such as armchair and table. 
Half the layouts are used as the training data, while the other half are used for validation and testing. The ratio of the validation and test splits is 1:9. mAP@$.5$, mAP@$.75$ and mAP@$[.5, .95]$ are used as evaluation metrics, where mAP@${*}$ represents the class-mean average precision with the intersection over union (IoU) threshold for counting as detected set to 50\% and 75\%, respectively. mAP@$[.5, .95]$ is the mean of the average precision for the IoU threshold between 0.50 and 0.95.

RendNet is applied for object detection in an R-CNN fashion \cite{girshick2014rich, girshick2015fast, ren2015faster}. Specifically, a proposal generation method is first used to generate candidate bounding boxes potentially containing objects. Then, the image region in each proposal is fed into RendNet to predict if the proposal is indeed an object as well as its object category. The same proposal generation method as YOLaT is used.  During training, given each proposal and its corresponding category or an extra category indicating that no object exists in the proposal, a cross-entropy loss is adopted to train RendNet. Note that the proposal generation method is based on VG format input and already generates precise bounding boxes for objects even with no extra offset regression branch \cite{jiang2021recognizing}.

\stitle{Experimental result.} For RG-based methods, we compare RendNet with the most popular object detection methods: one-stage methods including various variants of Yolov3~\cite{redmon2018yolov3}, Yolov4 \cite{bochkovskiy2020yolov4, wang2020scaled}, RetinaNet \cite{lin2017focal}, and two-stage methods including variants of Faster-RCNN with Pyramid Network (FPN)~\cite{lin2017feature}. For Yolov3, the -tiny variant is a smaller real-time model; the -spp variant uses Spatial Pyramid Pooling. For Yolov4, we apply a scaled Yolov4 \cite{wang2020scaled} with slightly more parameters and better performance. The Faster-RCNN-R$*$-FPN model series use backbone ResNets \cite{he2016deep} of different capacities, with R18, R34, and R50 in the model name standing for ResNet18, ResNet34, and ResNet50, respectively. Tab.~\ref{table:results-floorplan} shows that RendNet outperforms all baselines. 

We also compared our method with YOLaT \cite{jiang2021recognizing}, which is the first (and only, to the best of our knowledge) object detection method directly based on VG. Our method achieves comparable AP@.5 and significantly outperforms YOLaT in terms of AP@.75 by 3.6 percentage points. This result shows the effectiveness of leveraging merits from both RG and VG formats.

\subsection{3D object recognition}

%In this section 
We evaluate RendNet for 3D object recognition on the   FeatureNet~\cite{zhang2018featurenet} and Cluster3D~\cite{xiang2021cluster3d} datasets. 
%They are CAD models with labels about types or similarity. 

\subsubsection{3D object classification}

\begin{table}[t]
\centering
\begin{tabular}{lc}
\toprule
Model & Test acc. \\
\midrule
PointNet & 86.59 \\
PointNet++ & 95.78 \\
DGCNN & 96.15 \\
FeatureNet & 96.70 \\
\midrule
RendNet (Ours) & \textbf{99.31} \\
\bottomrule
\end{tabular}
\caption{\textbf{Classification accuracy (\%) on the FeatureNet dataset.} We outperform FeatureNet on their own dataset by a large margin.}
\label{tab:featurenetdb}
\end{table}

\begin{figure}[t]
    \centering
    \includegraphics[width=0.9\linewidth]{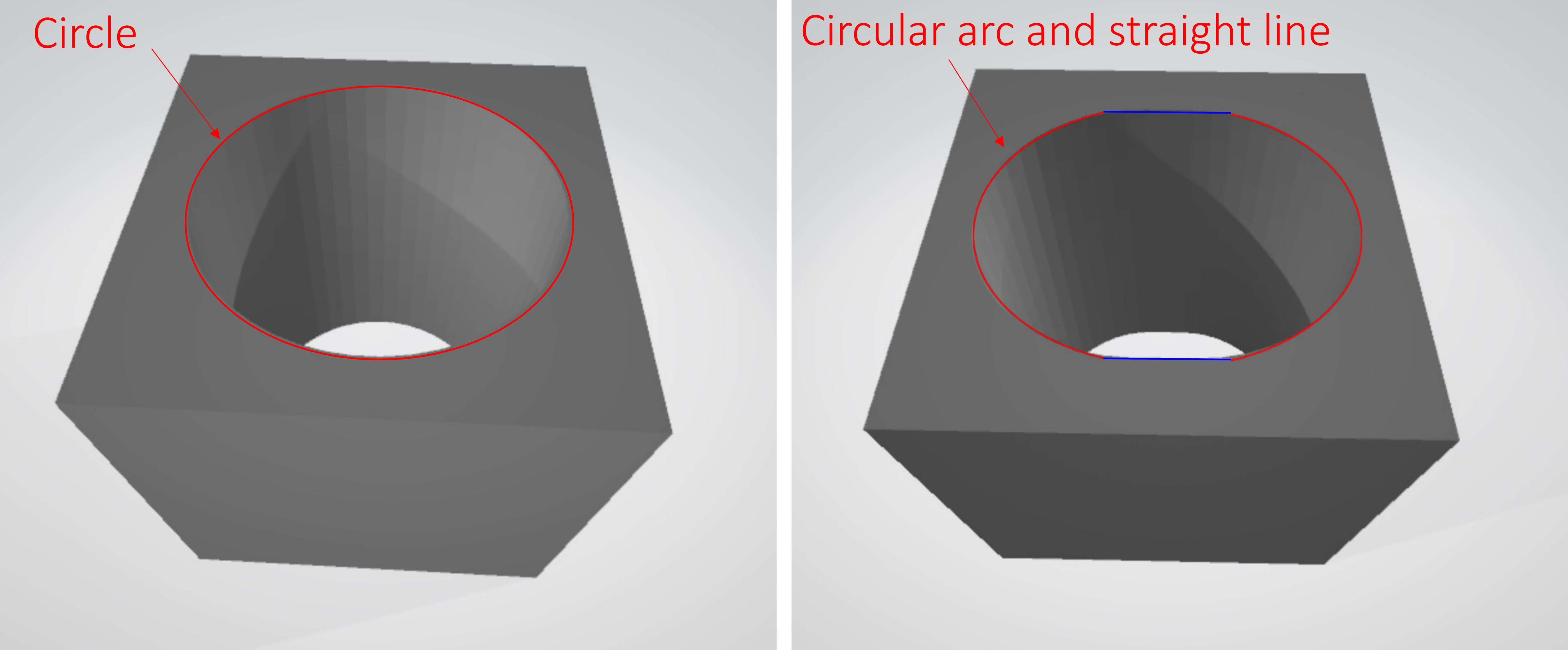}
    \caption{\textbf{Hard instances from the FeatureNet dataset.} The two instances are commonly misclassified by other methods, while RendNet handles them correctly. They belong to different classes: `through hole' (left) and `circular-end pocket' (right). }
    \label{fig:featnet_hardins}
\end{figure}

FeatureNet~\cite{zhang2018featurenet} has 24000 CAD models, consisting of 24 types of objects that contain different machining features. 
%The dataset is randomly split into 85\% for training and 15\% for testing. 
We randomly split the dataset where we utilize 85\% data for training and 
15\% data for testing.
%We trained each model in comparison for 100 epochs. 

The results for object classification are shown in Tab.~\ref{tab:featurenetdb}. We outperformed FeatureNet on their own dataset. Moreover, it can be seen that RendNet achieves superior performance on the dataset, with 99.31\% test accuracy while all other methods have errors greater than 3\%.

Two instances commonly misclassified by other methods are shown in Fig.~\ref{fig:featnet_hardins}. 
%They share the same topology and resemble each other in appearance, but belong to different machining features. 
They share the same topology (a cube with a hole) but belong to different machining features (the edge of the hole is a circle in the left instance while is composed of arcs and lines in the right instance). 
%Considering topology only there is no way to distinguish the two instances; considering the geometric appearance only it is likely to mistake one for another.
Considering topology only there is no way to distinguish the two instances. Based on the geometric appearance, it is hard for RG-based methods to classify them.
Our RendNet considers the local geometric features and the topology at the same time to distinguish these two instances correctly.
%By considering the local geometric features and the topology at the same time, however,  two instances can be distinguished correctly after noticing the straight lines connecting the two ends of the circular-end pocket on the second instance in Fig.~\ref{fig:featnet_hardins}.

With close interaction between the vector and raster streams, RendNet can effectively combine topology with geometric features, no matter it is planar 2D vector graphics or 3D models. Thus, it is able to benefit from both graphic formats. The design of RendNet works nicely with both 2D and 3D vector graphics.

\subsubsection{Similarity prediction for 3D objects}

% PointNet++: batch size=16x4GPUs, test batches=932, inference time = 5min54secs
% DGCNN: batch size=32x4GPUs, test batches=466, inference time = 4min18secs
% RendNet: batch size=80x4GPUs, test batches=187, inference time = 36secs
% PointNet: batch size=80x4GPUs, test batches=187, inference time = 25secs

Cluster3D~\cite{xiang2021cluster3d} is a dataset for non-categorical annotation of 3D CAD models. In Cluster3D, around 200000 similarity/dissimilarity pairs are annotated in a subset of the ABC dataset \cite{koch2019abc} that contains around 20000 CAD models. Note that according to the supplementary material of Cluster3D, the annotators are instructed to judge the similarity between models based on geometry and not functionality. Thus, topology does not account for so much in Cluster3D as other VG-based datasets, such as FeatureNet.

We partition the CAD models into two disjoint groups: the training group which contains 75\% of the models and the testing group which contains 25\% of the models. We train our RendNet with all pairwise annotations on the training group and test them on the testing group. For each pair of CAD models, we take the representations of two CAD models separately, concatenate them and feed them into a 2-layer MLP binary classifier to judge the similarity of the model pair. Experiments for inference time are conducted on an NVIDIA DGX with 4 V100 GPUs.

\begin{table}[t]
\centering
\begin{tabular}{lcc}
\toprule
Model & Test acc. & Inference time \\
\midrule
PointNet & 82.54 & 1.7 \\
PointNet++ & 84.19 & 23.7 \\
DGCNN & 85.47 & 17.3 \\
\midrule
RendNet (Ours) & \textbf{86.02} & 2.4 \\
\bottomrule
\end{tabular}
\caption{\textbf{Results on Cluster3D dataset.} Inference time is in milli-seconds per pair of models per V100 GPU, with batch size set to the limit of GPU memory.}
\label{tab:cl3d}
\end{table}

The results are shown in Tab.~\ref{tab:cl3d}. Since the dataset annotation biases against geometrical resemblance, the performance improvement upon previous methods is not as dramatic as in the previous detection and classification tasks. Nevertheless, the topological information from VG data can still boost model performance on Cluster3D. RendNet still achieves the best performance on Cluster3D.

Notably, the two popular methods on point clouds, PointNet++ and DGCNN, require around 8x-10x inference time compared to RendNet. For scanned scenes, PointNet++ spends 23.7 milliseconds per sample. When it comes to vast collections of CAD models, it is a bit too slow for PointNet++ to be applied under batch processing scenarios. Our RendNet only costs 36 seconds for validation per epoch in Cluster3D, compared to 6 minutes for PointNet++. 

\vspace{-6pt}
\subsection{Ablation study}

Here we validate our model design by running a classification task on the SESYD floor plan dataset. The train-test splitting scheme is the same as the detection task. We make small perturbations to the ground truth object boxes by cropping and extrusion, and take the resulting regions together with the labels for classification. % Tab.~\ref{tab:abl} reports the classification accuracy of different variants of RendNet under the setting.

\begin{table}[t]
\centering
\begin{tabular}{lc}
\toprule
Variant & Test error \\
\midrule
Raster stream only & 17.69 \\
Vector stream only & 2.80 \\
No graph edge features & 10.68 \\
No final block & 1.99 \\
Ensemble of VG/PointNet & 1.70 \\
\midrule
Full RendNet & \textbf{0.81} \\
\bottomrule
\end{tabular}
\caption{\textbf{Ablation study on components of RendNet.} Different variants are evaluated on the SESYD floor plan classification task.}
\label{tab:abl}
\end{table}

\stitle{Two streams in residual blocks.} We zero out the outputs from vector or raster stream in the residual blocks in this experiment (`raster/vector stream only' entries in Tab.~\ref{tab:abl}). It can be seen that both streams in the residual blocks contribute to the power of RendNet. The vector stream not only processes local node embeddings but is also critical for propagating information between different parts of the object, while the raster stream only handles local neighborhoods of nodes. Thus without the vector stream, the overall performance drops more than without the raster stream.

\stitle{Global feature aggregation block.} We remove the final global feature aggregation block in RendNet and use global max pooling on hypergraph nodes instead (`no final block' entry in Tab.~\ref{tab:abl}). This final block enables RendNet to view the whole rasterization result at the end, which leads to a more powerful global feature aggregation process.

\stitle{Graph edge features.} We input as edge features the type and starting/ending direction vector of curve segments to RendNet. In this experiment we replace the vector graphic module with a bare graph convolution \cite{DBLP:conf/iclr/KipfW17} that does not take in edge features (`no graph edge features' entry in Tab.~\ref{tab:abl}). It shows a significant performance decrease. The edge features are important for vector graphics recognition since topology is also decided by the type and the curvature of curves, besides connectivity between nodes.

\stitle{Latent space rasterization.} In this experiment, we study the effectiveness of incorporating the rendering process into the model. We zero out the outputs from the raster stream in the residual blocks and aggregate the output of the final residual block with a direct graph max-readout operation. In addition, we create an independent stream that directly applies a PointNet to the rasterized point cloud. The two representations (from VG and PointNet, respectively) are concatenated and fed into the MLP classifier. The result is shown in `ensemble of VG/PointNet' entry in Tab.~\ref{tab:abl}. The test error is lower than using VG or RG only, indicating that the model can benefit from both modalities. Yet the error is still $\sim100\%$ higher relative to the original RendNet -- incorporating the rendering process is much more powerful than a simple combination of VG and RG inputs.

\subsection{Qualitative results of RendNet}

\begin{figure}
    \centering
    \includegraphics[width=0.85\linewidth]{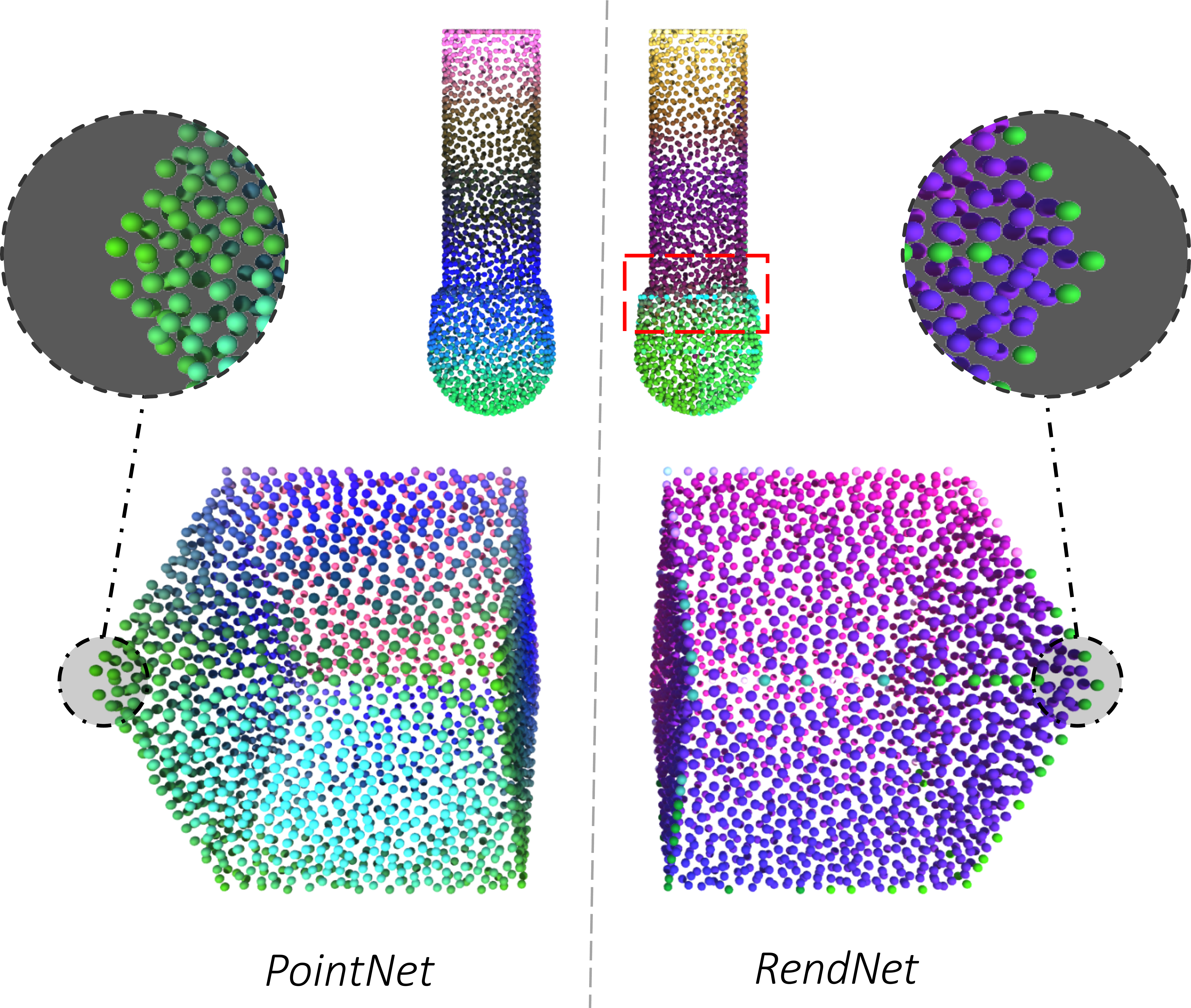}
    \caption{\textbf{Comparison between representations from RendNet and PointNet.} RendNet captures finer structures.}
    \label{fig:featvis}
    \vspace{-6pt}
\end{figure}

To investigate the representations learnt by RendNet, we trained a RendNet in an unsupervised manner with BYOL \cite{grill2020bootstrap} on the ABC \cite{koch2019abc} dataset. A PointNet \cite{qi2017pointnet} is also trained under the same settings for comparison. We visualized the final point-wise representations by performing a Principal Component Analysis (PCA) to 3 dimensions and mapping them to RGB color components. The results are shown in Fig.~\ref{fig:featvis}. The upper object is a bolt. There is a fast change of representation for RendNet marked by the red box. Our representation differentiates the head and body of the bolt clearly. For the cubic object, both PointNet and RendNet emphasize edges and corners, but RendNet captures much sharper edges than PointNet. This means that RendNet is able to capture semantics more accurately.

\section{Conclusion}
In the paper, we propose RendNet, which leverages vector graphics and raster graphics to recognize 2D and 3D objects. We also design a new rendering method in latent space. Various experiments on 2D and 3D object recognition demonstrate that RendNet has higher performance and good efficiency compared with baselines. In the future, we can integrate more techniques to further improve the performance, such as pre-training on large VG datasets.
% It helps to learn the primitive representations based on the graph structure and the Euclidean space in a hybrid fusion.

%%%%%%%%% REFERENCES
{\small
\bibliographystyle{ieee_fullname}
\bibliography{egbib}
}

\end{document}